\documentclass[conference]{IEEEtran}
\usepackage{graphicx}
\usepackage{amsmath}
\usepackage{hyperref}
\usepackage{float}
\usepackage{placeins}
\usepackage{orcidlink}

\title{Synthetic Cardiac MRI Image Generation using Deep Generative Models}

\author{
  \IEEEauthorblockN{Ishan Kumarasinghe\orcidlink{0009-0004-3110-1290}}
  \IEEEauthorblockA{
    \textit{Department of Computer Engineering} \\
    \textit{University of Peradeniya}\\
    Peradeniya, Sri Lanka \\
    \texttt{e20211@eng.pdn.ac.lk}
  }
  \and
  \IEEEauthorblockN{Dasuni Kawya\orcidlink{0009-0006-2921-4573}}
  \IEEEauthorblockA{
    \textit{Department of Computer Engineering} \\
    \textit{University of Peradeniya}\\
    Peradeniya, Sri Lanka \\
    \texttt{e20197@eng.pdn.ac.lk}
  }
  \and
  \IEEEauthorblockN{Madhura Edirisooriya\orcidlink{0009-0007-8166-3736}}
  \IEEEauthorblockA{
    \textit{Department of Computer Engineering} \\
    \textit{University of Peradeniya}\\
    Peradeniya, Sri Lanka \\
    \texttt{e20093@eng.pdn.ac.lk}
  }

  \and
  \IEEEauthorblockN{Isuri Devindi\orcidlink{0009-0005-6615-7937}}
  \IEEEauthorblockA{
    \textit{Department of Computer Engineering} \\
    \textit{University of Maryland}\\
    College Park, USA \\
    \texttt{isurid@umd.edu}
  }

  \and
  \IEEEauthorblockN{Isuru Nawinne\orcidlink{0009-0001-4760-3533}}
  \IEEEauthorblockA{
    \textit{Department of Computer Engineering} \\
    \textit{University of Peradeniya}\\
    Peradeniya, Sri Lanka \\
    \texttt{isurunawinne@eng.pdn.ac.lk}
  }

  \and
  \IEEEauthorblockN{Vajira Thambawita\orcidlink{0000-0001-6026-0929}}
  \IEEEauthorblockA{
    \textit{Department of Holistic Systems} \\
    \textit{SimulaMet}\\
    Oslo, Norway \\
    \texttt{vajira@simula.no}
}
}

\begin{document}
\maketitle

\begin{abstract}
    Synthetic cardiac MRI (CMRI) generation has emerged as a promising strategy to overcome the scarcity of annotated medical imaging data. Recent advances in GANs, VAEs, diffusion probabilistic models, and flow-matching techniques aim to generate anatomically accurate images while addressing challenges such as limited labeled datasets, vendor variability, and risks of privacy leakage through model memorization. Mask-conditioned generation improves structural fidelity by guiding synthesis with segmentation maps, while diffusion and flow-matching models offer strong boundary preservation and efficient deterministic transformations. Cross-domain generalization is further supported through vendor-style conditioning and preprocessing steps like intensity normalization. To ensure privacy, studies increasingly incorporate membership inference attacks, nearest-neighbor analyses, and differential privacy mechanisms. Utility evaluations commonly measure downstream segmentation performance, with evidence showing that anatomically constrained synthetic data can enhance accuracy and robustness across multi-vendor settings. This review aims to compare existing CMRI generation approaches through the lenses of fidelity, utility, and privacy, highlighting current limitations and the need for integrated, evaluation-driven frameworks for reliable clinical workflows.
\end{abstract}

\begin{IEEEkeywords}
Cardiac MRI, Synthetic Data, Generative Models, Fidelity, Privacy
\end{IEEEkeywords}

\section{INTRODUCTION}
Synthetic cardiac MRI (CMR) generation has gained significant attention as a promising solution to the longstanding problem of limited annotated medical imaging data \cite{Diller2020}, \cite{Jaen-Lorites2022}. Manual segmentation of cardiac structures is time-consuming, requires expert clinicians, and remains difficult to scale, particularly for multi-frame cine MRI, where numerous slices and temporal phases must be labeled \cite{Amirrajab2022}. Alongside these constraints, strict privacy regulations, the rarity of certain cardiac pathologies, and substantial variation in acquisition protocols across scanner vendors further restrict the availability of high-quality, diverse training datasets \cite{Diller2020}. These challenges collectively motivate the development of generative models capable of producing realistic, clinically meaningful synthetic CMR data under limited labeled data settings \cite{Diller2020}\cite{Amirrajab2022}.\\

Generative modelling techniques such as generative adversarial networks (GANs), variational autoencoders (VAEs), diffusion probabilistic models, and more recent flow-matching frameworks have all demonstrated varying degrees of success in capturing cardiac anatomy \cite{Liu2024}. While GANs offer visually plausible results, they often suffer from instability and artifacts \cite{Zhang2025}; VAEs provide smooth latent representations but may lack fine structural detail \cite{Jaen-Lorites2022}. Diffusion models, through their iterative noise-removal process, have recently emerged as state-of-the-art for high-fidelity medical image synthesis\cite{rombach2022ldm}. Flow-matching models present a newer alternative that models continuous transformations\cite{Moschetto2025} between distributions, but their performance in mask-conditioned cardiac MRI generation remains under-explored.\\\\
Despite these advances, much of the existing work still relies on unconditional synthesis or simplistic augmentation strategies\cite{Zhang2025}. Such approaches fail to ensure anatomical precision, particularly regarding subtle structural variations \cite{Strzelecki2025} in the left ventricle (LV), right ventricle (RV), and myocardium (MYO). Conditioning on segmentation masks presents a more promising direction by explicitly guiding generative models to align with cardiac structure\cite{Amirrajab2022}\cite{AlKhalil2023}\cite{Heo2025}. However, only a limited number of studies have rigorously evaluated conditional generative models across the three critical dimensions required for clinical usefulness: fidelity, downstream task utility\cite{Xing2023}, and privacy.\\\\
Privacy concerns remain especially relevant in medical imaging. Because cardiac MRIs inherently reflect individual anatomical characteristics, synthetic images may unintentionally reveal patient identity if generative models memorize training examples~\cite{bergen}. Privacy vulnerabilities such as membership inference attacks and nearest-neighbour similarity analysis reveal the need for explicit privacy assessments\cite{bergen}. Nevertheless, privacy evaluation is often absent or only superficially addressed in prior work \cite{Liu2024}. Achieving a balance between generating realistic, clinically useful images and ensuring strong privacy protection therefore represents a core challenge\cite{Xing2023}.\\\\
Multi-vendor heterogeneity adds another layer of complexity\cite{AlKhalil2023}. Differences in scanner manufacturers, acquisition sequences, and reconstruction algorithms introduce substantial domain shifts. Models trained on a single vendor frequently degrade when tested on images from unseen scanners\cite{Campello2021}. Synthetic augmentation has the potential to mitigate this by enriching training datasets with diverse, vendor-variant image styles, but only if the generative models are carefully validated through downstream tasks such as segmentation\cite{AlKhalil2023}.\\\\
To navigate these challenges, a structured assessment of conditional generative models, with emphasis on diffusion and flow-matching techniques, is required. This review therefore examines existing CMRI synthesis approaches through the lenses of fidelity, utility, and privacy to identify methodological strengths, expose current gaps, and inform the development of dependable, privacy-conscious synthetic data frameworks for downstream clinical applications.

\subsection{Background and Clinical Importance of Cardiac MRI}
\subsubsection{Cardiac MRI in Clinical Diagnostics}
Cardiac magnetic resonance imaging (MRI) is a key modality for diagnosing and monitoring cardiovascular diseases because it provides high-resolution, high-contrast visualization of cardiac structure and function \cite{Gheorghi2022}\cite{Strzelecki2025}. Its ability to delineate the left and right ventricles, myocardium, and atrial chambers enables accurate assessment of chamber volumes, myocardial mass, wall motion, and ejection fraction, critical biomarkers in conditions such as ischemic heart disease, cardiomyopathies, valvular disorders, and congenital abnormalities\cite{Oscanoa2023}.
Clinical workflows benefit from a range of imaging sequences, each capturing different aspects of myocardial health. Cine MRI provides dynamic functional assessment across the cardiac cycle\cite{Li2025}; late gadolinium enhancement (LGE) reveals scar or fibrosis; and T1/T2 mapping quantifies tissue composition changes relevant to inflammation or edema\cite{Oscanoa2023}. Together, these sequences form comprehensive diagnostic protocols. However, extracting quantitative measures typically requires manual contouring by experts, which is time-consuming and limits scalability\cite{Campello2021}\cite{Gheorghi2022}. \\

\subsubsection{Advantages Over Other Imaging Modalities}
Cardiac MRI (CMR) stands out among cardiac imaging techniques because it provides high-resolution, high-contrast images without ionizing radiation\cite{Al-Haidri2023}\cite{Liu2024}, unlike CT, making it ideal for repeated acquisitions and safe long-term follow-up. Compared to echocardiography, CMR offers consistent image quality that does not depend on operator skill or patient anatomy, ensuring reliable datasets for machine learning\cite{Abdusalomov2023}.
A major advantage for generative modelling is CMR’s ability to produce multiple standardized sequences, such as cine, T1w, T2w, and LGE\cite{Liu2024}, each capturing unique structural or functional information. This variety enables models to learn richer anatomical and tissue characteristics, supporting high-fidelity synthetic image generation\cite{Amirrajab2022}.
MRI also provides flexible multi-planar acquisitions\cite{Strzelecki2025} with excellent anatomical coverage, enabling consistent short-axis, long-axis, and 4D cine sequences\cite{Li2025}. For generative modelling, this flexibility means that models can learn robust anatomical structures across varying slice orientations, important for 2D and 3D synthetic reconstruction tasks\cite{AlKhalil2023}. The clarity of myocardial borders in MRI improves segmentation ground truth quality, supporting more accurate supervised learning and ensuring synthetic images encode precise anatomical boundaries.
Although multi-vendor variability introduces challenges, this heterogeneity benefits synthetic data research by allowing models to learn cross-domain variations and generalize better to real-world clinical environments\cite{AlKhalil2023}.
These combined strengths make CMR the most suitable modality for mask-conditioned synthetic generation, fidelity-utility evaluation, and privacy-aware modelling in cardiac imaging.

\subsection{Challenges in Cardiac MRI Data Availability and Usage} 
\subsubsection{Limited Annotated Data}
The scarcity of annotated cardiac MRI datasets remains one of the central limitations for both clinical AI and synthetic data research\cite{Liu2024}. High-quality segmentation labels require time-consuming manual effort from expert clinicians\cite{AlKhalil2023}, often across dozens of slices and multiple cardiac phases\cite{Campello2021}. This workload makes it difficult to build large, diverse labeled datasets, especially for 4D cine MRI\cite{Oscanoa2023}. As a result, models are frequently trained on small samples that may not capture the full variability of cardiac anatomy, scanner vendors, or rare pathologies.\\\\
Privacy regulations such as GDPR further restrict multi-institutional data sharing\cite{delCastillo2025}, creating fragmented data silos that cannot be easily pooled\cite{Diller2020}. This exacerbates overfitting risks~\cite{Liu2024} and limits the representativeness of training, validation, and test sets. 
Overall, limited annotated data is not only a practical constraint but a methodological challenge influencing model design, evaluation, and privacy. Addressing this requires hybrid pipelines that maximize unlabeled data usage, incorporate synthetic augmentation, and include robust privacy assessments to prevent patient-specific leakage.\\
\subsubsection{Vendor and Scanner Variability}
Multi-center studies reveal significant performance degradation when models trained on one dataset are deployed on data from different scanner vendors, imaging protocols, or acquisition sites\cite{AlKhalil2023}\cite{Campello2021}. Even standardized cine sequences can differ subtly across vendors, causing models to misinterpret vendor-specific characteristics as anatomical changes\cite{Campello2021}. The M\&Ms (Multi-Center, Multi-Vendor, Multi-Disease) challenge demonstrated that models achieving 93\% Dice score on single-center data experience performance drops to 88\% when tested on multi-center, multi-vendor datasets due to variations in image intensity, contrast, spatial resolution, and noise characteristics\cite{Campello2021}.

\subsection{Motivation for Synthetic Data Generation in Medical Imaging}
Synthetic data generation offers a promising solution to limited annotated datasets and strict privacy regulations by: 

\begin{enumerate}
    \item \textbf augmenting limited real datasets with diverse, anatomically plausible examples\cite{Xing2023}.
    \item \textbf simulating rare pathologies underrepresented in training data\cite{delCastillo2025}\cite{Amirrajab2023}.
    \item \textbf enabling cross-vendor domain adaptation without requiring additional acquisitions\cite{AlKhalil2023}.
    \item \textbf potentially preserving privacy by sharing trained generative models rather than patient data\cite{delCastillo2025}.
\end{enumerate}
Traditional data augmentation techniques, including geometric transformations (rotation, scaling, flipping) and intensity modifications (brightness, contrast adjustment) provide only modest improvements and cannot introduce novel anatomical variations\cite{Strzelecki2025}.

\section{Preprocessing Strategies in Cardiac MRI}
Pre-processing profoundly influences synthetic image fidelity, cross-vendor generalization, and training stability. Preprocessing decisions represent the first critical control point for downstream generative model performance\cite{Mourad2024}.

\subsection{Rescaling and Intensity Normalization}
Rescaling and intensity normalization are essential preprocessing steps in cardiac MRI pipelines, ensuring consistent input quality across diverse scanners, vendors, and acquisition protocols\cite{AlKhalil2023}. Because cardiac MRIs vary in voxel size, field of view, and slice count, spatial rescaling aligns all scans to a fixed resolution and geometry\cite{Gheorghi2022}, enabling generative and segmentation models to operate on uniform inputs. Care is taken to preserve anatomical proportions typically by maintaining aspect ratio and using padding to avoid distortions that would affect mask-conditioned synthesis or downstream clinical measurements such as ventricular volume.
Intensity normalization addresses the lack of an absolute MRI intensity scale. Common approaches include:

\begin{enumerate}
    \item \textbf{Min-max normalization} - Rescaling intensity to [0,1] or [-1,1] range used extensively in cardiac synthesis studies\cite{Li2025}.
    \item \textbf{Z-score normalization} - Computing mean and standard deviation across foreground pixels, normalizing to zero-mean, unit-variance distributions\cite{Amirrajab2022}\cite{Strzelecki2025}.
    \item \textbf{Percentile clipping} - Clipping intensities to 2nd-98th percentiles before normalization reduces outlier artifacts from motion or flow artifacts\cite{AlKhalil2023}.
    \item \textbf{N4ITK bias field correction} - Correcting spatially varying signal inhomogeneities persistent across CMR sequences\cite{Moschetto2025}\cite{Strzelecki2025} .
\end{enumerate}

\subsection{Cropping and ROI Extraction}
Region-of-interest extraction reduces computational burden and improves model focus:

\begin{enumerate}
    \item \textbf Bounding box cropping: Automatically detecting cardiac regions using CNNs, then cropping to 128×128 or 256×256 pixels centered on the heart\cite{AlKhalil2023}.
    \item \textbf Consistent field-of-view: Standardizing FOV across subjects by resampling\cite{AlKhalil2023}.
\end{enumerate}
\subsection{Slice Selection and Temporal Frames}
CMR acquisitions contain multiple spatial and temporal dimensions\cite{Oscanoa2023} requiring careful selection:
\begin{enumerate}
    \item \textbf{End-diastolic (ED) and end-systolic (ES) frames} - Most clinical segmentation studies use ED and ES frames representing maximum and minimum ventricular volume states\cite{Campello2021}.
    \item \textbf{Short-axis (SAX) stack} - Standard acquisition spanning apex to base with 8–10 mm slice thickness and variable inter-slice distance (typically 6-10 mm)\cite{Gheorghi2022}.
    \item \textbf{Central slice selection} - Studies commonly select 2-4 central slices per volume to focus on regions with clear left ventricle (LV), right ventricle (RV), and myocardium (MYO) visibility\cite{Torfi2021}.
\end{enumerate}
Frame selection directly impacts synthesizing model training: including both ED and ES improves robustness to temporal variability, and including unlabeled apical/basal slices struggles with synthesizing plausible appearance of anatomy\cite{AlKhalil2023}.

\section{Synthetic Medical Image Generation}
\begin{figure}
    \centering
    \includegraphics[width=1\linewidth]{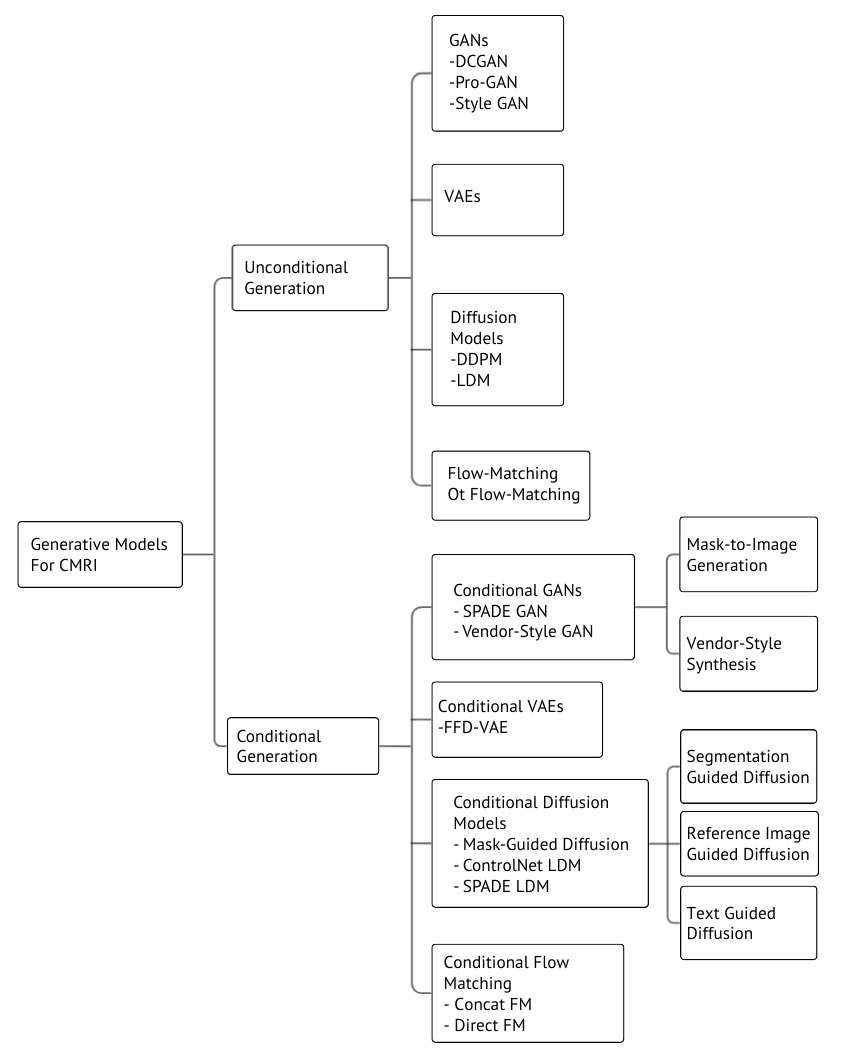}
    \caption{Taxonomy of generative models used in CMRI synthesis}
    \label{fig:modelsoverview}
\end{figure}

\textbf{Figure}~\ref{fig:modelsoverview} provides an overview of the major generative model families used for CMRI synthesis, highlighting both unconditional and conditional architectures that are discussed in this section.

\subsection{Unconditional Generative Models}
Unconditional generative models learn to map random noise vectors directly to realistic images without any auxiliary conditioning variables, modeling the overall data distribution rather than specific anatomical or pathological subsets \cite{Diller2020}\cite{Amirrajab2022}. In cardiac imaging, unconditional models primarily serve to expand datasets and investigate general appearance variability \cite{Diller2020}.
Unconditional generation provides baseline capability but lacks anatomical control. GANs learn mappings from noise vectors to images through adversarial training\cite{Amirrajab2022}. VAEs learn lower-dimensional latent representations enabling diverse generation\cite{Kebaili2023}. Diffusion models reverse learned noise processes through iterative denoising, achieving superior fidelity and diversity\cite{rombach2022ldm}.\\

\subsubsection{Unconditional GAN Models}
Unconditional generative adversarial networks (GANs) learn to map random noise vectors to realistic images by training a generator and discriminator in an adversarial framework, without requiring auxiliary conditioning information such as class labels or segmentation masks \cite{Alrashedy2022}\cite{bergen}\cite{Kebaili2023}. Early unconditional GAN architectures, including DCGAN and Progressive GAN (ProGAN), demonstrated the ability to synthesize plausible medical images by progressively increasing resolution during training, thereby improving stability and output quality \cite{Diller2020}\cite{Skandarani2021}\cite{Amirrajab2022}. In cardiac imaging, unconditional GANs have been applied to generate synthetic MRI frames for data augmentation purposes, ProGAN-generated cardiac MR images can augment training sets for congenital heart disease applications. Unconditional GAN frameworks were used for brain MRI synthesis, highlighting their potential for expanding limited datasets\cite{Diller2020}\cite{Alrashedy2022}.\\

However, unconditional GANs exhibit notable limitations: they lack explicit control over anatomical structures or pathology, often suffer from mode collapse, and may produce images with unrealistic artifacts \cite{Kebaili2023}\cite{delCastillo2025}.\\

\subsubsection{Unconditional Diffusion Models}
Unconditional diffusion models have emerged as a robust alternative to GAN and VAE-based synthesis for medical imaging, offering improved stability and fidelity when modeling complex anatomical structures. These models operate by reversing a predefined forward Markov process that gradually corrupts data with Gaussian noise, enabling high-quality image generation through iterative denoising steps \cite{Strzelecki2025}. Unlike GANs, which depend on adversarial optimization and are susceptible to mode collapse, diffusion models benefit from a well-posed likelihood-based objective that facilitates broader distribution coverage and more reliable convergence \cite{Strzelecki2025}. Their capacity to reconstruct images through progressive refinement has fueled widespread adoption across medical imaging tasks, with recent studies demonstrating high-fidelity cardiac textures and ventricular structures synthesized using Denoising Diffusion Probabilistic Models (DDPMs) implemented within frameworks such as MONAI \cite{Kebaili2023}\cite{Urcia-Vzquez2024}.\\

Despite their strengths in realism and diversity, unconditional diffusion models face practical challenges. They require substantial computational resources due to the multi-step sampling process, though recent work exploring random latent embeddings has shown promise in accelerating inference without compromising quality \cite{Hejrati2025}. Furthermore, privacy concerns persist, as nearest-neighbor leakage can occur when unconstrained diffusion models inadvertently reproduce features of specific training samples, prompting the need for formal safeguards such as differential privacy mechanisms \cite{Xing2023}\cite{Torfi2021}.\\

\subsubsection{Unconditional Flow Matching Models}
Flow matching is an emerging class of generative models that learns continuous normalizing flows by regressing a velocity field to transform a simple prior distribution (e.g., Gaussian noise) into the target data distribution \cite{Yazdani2025}. Unlike diffusion models, which require hundreds of denoising steps, flow matching models directly learn a transport map, enabling significantly faster sampling while maintaining competitive image quality. Flow matching bridges the gap between speed and quality in medical image synthesis, achieving inference times substantially lower than traditional diffusion approaches without sacrificing fidelity \cite{Yazdani2025}\cite{Moschetto2025}\cite{Hadzic2025}.\\

In unconditional settings, flow matching models generate images by sampling from noise and following the learned velocity field trajectory to produce realistic outputs . Applying flow matching for MRI-CT and CBCT-CT image synthesis, showing that unconditional flow-based generation can produce anatomically coherent medical images \cite{Hadzic2025}. Flow matching achieves intermediate performance between GANs and diffusion models, offering faster inference than diffusion while providing better diversity than GANs. \\ 

However, unconditional flow matching lacks explicit anatomical control, limiting its applicability to pathology-specific or mask-conditioned generation tasks. This motivates the exploration of conditional flow matching frameworks for cardiac MRI synthesis \cite{Yazdani2025}\cite{Moschetto2025}.\\

\subsection{Conditional Generative Models}
Conditional generative models extend unconditional models by incorporating additional information such as labels, segmentation masks, or text to control synthesis, enabling anatomy- or phenotype-specific image generation \cite{Amirrajab2023}\cite{Amirrajab2022}\cite{Kebaili2023}. Recent advancements have moved beyond simple class labels, employing high-dimensional spatial masks to enforce anatomical consistency \cite{Li2025}\cite{Konz2024}. In cardiac imaging, conditional models improve fidelity, diversity, and clinical relevance of synthetic MR data for augmentation and downstream tasks \cite{Amirrajab2023}\cite{Amirrajab2022}\cite{AlKhalil2023}. Furthermore, the architectural landscape has evolved from Conditional Generative Adversarial Networks (cGANs), which require careful balancing of generator-discriminator dynamics to Conditional Diffusion Models, which offer stable training objectives and fine-grained control over the synthesis process through mechanisms like classifier guidance and cross-attention \cite{Dhariwal2021}. The following sections review these models, categorized by their conditioning strategies and architectures.\\

\subsubsection{Deep Convolutional Generative Adversarial Network (DCGAN)}
A DCGAN architecture was implemented as a core component of proposed models for synthesizing medical images. Although the DCGAN concept is general, the network utilizes Convolutional Neural Networks (CNNs) with specific filters to detect and reproduce essential patterns from the data distribution. The generated images, intended to mimic sagittal plane brain MRIs, were successfully used to assess the efficacy of downstream classification tasks \cite{Mourad2024}.\\

\subsubsection{Conditional Generative Adversarial Network (Vendor-Specific Synthesis)}
Conditional GANs were implemented to facilitate domain adaptation in cardiac MR segmentation by synthesizing data in a vendor-specific style. The model involved training two separate generators for different scanner vendors (Vendor A and Vendor B). This allowed the system to synthesize images with the appearance of one vendor based on anatomical labels acquired from another, effectively utilizing conditional synthesis to address domain shift \cite{AlKhalil2023}.\\

\subsubsection{GauGAN (SPADE-based Architecture)}
A generative approach based on the GauGAN architecture was implemented to synthesize short-axis cardiac MR stacks. This model utilized a high-level conditioning mechanism to control the generated images, enabling the synthesis of data with a wide and dense distribution of cardiac functional metrics, such as ejection fraction (EF). This synthetic data was subsequently used for pre-training a regression network for cardiac function quantification \cite{Gheorghi2022}.\\

\subsubsection{Reconstruction Conditional GAN (RecCGAN)}
The RecCGAN framework was implemented for cardiac MR under-sampled image reconstruction. This specific Conditional GAN architecture integrates a hybrid spatial and frequency loss function to calculate the difference between the target and reconstructed images. The CGAN objective is supplemented with an L1 distance to enhance accuracy and reduce image blurring during the reconstruction process \cite{Al-Haidri2023}.\\

\subsubsection{Wasserstein Generative Adversarial Network (Cardiac Biasing)}
A conditional generative model featuring a conditional biasing mechanism was implemented for cardiac imaging synthesis. This model employed residual and attentional blocks along with a sinusoidal embedding to transform scalar covariates (e.g., age or BMI) into vector conditions. This structure enabled the fine-grained control necessary to induce plausible morphological variations, such as synthesizing images representing cardiac aging or modifying the apparent BMI of subjects \cite{Skorupko2025}.\\

\subsubsection{Rényi Differential Privacy-Conditional GAN (RDP-CGAN)}
The RDP-CGAN framework was proposed and implemented to ensure Rényi Differential Privacy (RDP) during synthetic data generation, especially for tabular/sequential medical records. This model uses a Convolutional GAN to capture temporal and correlated information, and crucially, enforces RDP exclusively on the discriminator training while utilizing a pre-trained convolutional autoencoder for handling mixed-type data features \cite{Torfi2021}.\\

\subsubsection{Feature Factorization and Distillation VAE (FFD-VAE)}
The Feature Factorization and Distillation (FFD-VAE) architecture was implemented to enhance cross-domain generalization. This VAE employs two separate encoders to disentangle the latent space, learning independent representations for segmentation knowledge (z iseg) and domain-related knowledge (z dom). This factorization strategy ensures that the essential anatomical structure is preserved while allowing the extracted domain features to be used as a condition for subsequent diffusion models \cite{Zhang2025}.\\

\subsubsection{Segmentation-Guided Diffusion Model}
A Segmentation-Guided Diffusion Model was implemented to condition image generation on a multi-class anatomical mask (m), aiming to sample from the distribution p(x/m). This model demonstrated high faithfulness, accurately reproducing anatomical realism required for downstream tasks, like training segmentation networks, with performance nearly matching models trained purely on real data \cite{Konz2024}.\\

\subsubsection{Source Domain-Trained Diffusion Model (SD-DM)}
The Source Domain-Trained Diffusion Model (SD-DM) was implemented to mitigate domain shift in cardiac MR imaging by performing a reference-based adaptation. The sampling process is iteratively refined, conditioned on a reference image, to ensure the synthetic data maintains structural consistency with anatomical features while adopting the visual characteristics of the source training domain \cite{Wong2025}.\\

\subsubsection{Controllable Mask Diffusion Framework}
A multi-label conditioned diffusion framework was implemented, involving several complex steps to generate pseudo-samples. The process includes training a pseudo-label mask diffusion model, followed by a cardiac MR image implicit diffusion model. This second stage integrates the SPADE module into its decoding process to enforce pixel-level alignment with the pseudo-labels, coupled with a label screening strategy to filter noisy inputs \cite{Li2025}.\\

\subsubsection{Latent Diffusion Model (Pathology and Modality Conditioning)}
A Latent Diffusion Model (LDM) was implemented for conditioned brain MRI generation. The model generates images in a compressed latent space and is specifically conditioned on auxiliary inputs specifying both pathology (e.g., Glioblastoma, Sclerosis) and the target acquisition modality (e.g., T1w, T2w). This allows the model to generate synthetic images representing configurations not explicitly present in the training data, demonstrating extrapolation capability \cite{delCastillo2025}.\\

\subsubsection{ControlNet (Multi-Conditional LDM)}
ControlNet, built upon a Stable Diffusion 2.1 base model, was implemented and fine-tuned on UK Biobank cardiac MR data. This model allows for multi-modal conditioning using both text prompts and segmentation masks, leveraging the LDM's cross-attention mechanisms for flexible control over synthesis, and demonstrating that complex conditional diffusion can be achieved with modest computational resources \cite{Skorupko2025}.\\

\subsubsection{MOTFM (Medical Optimal Transport Flow Matching)}
The Medical Optimal Transport Flow Matching (MOTFM) framework was implemented for efficient, high-quality medical image synthesis, particularly for echocardiographic images. This model uses a UNet backbone enhanced with attention layers to estimate the optimal velocity field. MOTFM demonstrated superior efficiency, achieving performance comparable to 50-step DDPM in significantly fewer steps (e.g., 10 steps) for mask-conditioned synthesis \cite{Yazdani2025}.\\

\subsubsection{Conditional Flow Matching (Concat. FM and Direct FM)}
Conditional Flow Matching (CFM) was implemented with two distinct conditioning strategies for T1w-to-T2w MRI translation. Concat. FM conditions the flow by concatenating the T1w slice with the intermediate flow sample. Direct FM utilizes the T1w slice directly as the source sample (z0). Both strategies train the model to match the constant conditional velocity field (z1-z0) to learn the flow dynamics \cite{Moschetto2025}.\\

\section{Fidelity Evaluation in Generative Models}
The correct evaluation of synthetic medical images is crucial, as the quality of these images can impact human life and health \cite{Abdusalomov2023}. Since medical imaging modalities, such as MRI, CT, and Echo images, are typically noisier, blurrier, and possess less defined edges compared to common images (e.g. cats or faces), specialized and accurate evaluation metrics are necessary \cite{Abdusalomov2023}. While metrics like Mean Absolute Error (MAE), Mean Squared Error (MSE), and Structural Similarity Index Measure (SSIM) are traditionally used, they may not always accurately reflect the true visual quality of the medical image \cite{Kebaili2023}.\\

Fidelity generally refers to the realism of the synthetic image, quantified by how closely its feature distribution and structural characteristics align with those of real data \cite{Xing2023}\cite{Sun2023}. The Fréchet Inception Distance (FID) is widely adopted as the de facto standard metric for overall sample quality in generative models because it captures aspects of both fidelity and diversity \cite{Dhariwal2021}. A lower FID score indicates that the synthetic images' feature distribution is closer to that of the real images \cite{Sahoo2021}. In addition to FID, quantitative measures used frequently in cardiac MRI synthesis include Structural Similarity Index Measure (SSIM), Peak Signal-to-Noise Ratio (PSNR), and its multi-scale variant, MS-SSIM \cite{Strzelecki2025}\cite{Cheng2024}\cite{Oscanoa2023}. For cardiac images specifically, assessing fidelity often goes beyond simple pixel comparisons to include measuring clinically relevant anatomical parameters, such as the Left Ventricle (LV)/Myocardium (Myo) Ratio, Right Ventricle (RV)/LV Ratio, and Myocardium Thickness, ensuring the synthetic data maintains structural integrity \cite{Strzelecki2025}.\\

However, reliance on distribution-based metrics such as the FID alone is insufficient for evaluating synthetic medical images. A low FID score may still occur in cases of mode collapse or overfitting, where the model generates near-identical replicas of training samples rather than truly diverse images \cite{Abdusalomov2023}. For this reason, application-driven utility evaluation is widely regarded as the most reliable assessment strategy \cite{Kebaili2023}\cite{AlKhalil2023}\cite{Amirrajab2022}.\\

In terms of generative model comparison, recent work indicates that Diffusion Models often achieve superior fidelity, resulting in the lowest FID scores during unconditional image generation compared to GANs and VAEs \cite{Strzelecki2025}\cite{bubeck2022}. Conversely, for image-to-image translation tasks (e.g., T1w-to-T2w MRI translation), GAN-based methods like Pix2Pix have sometimes demonstrated superior structural fidelity and computational efficiency over Diffusion Models and Flow Matching techniques \cite{Moschetto2025}. This complexity highlights the ongoing challenge of navigating the inevitable trade-off that exists among image fidelity, sample variety, and data privacy \cite{Xing2023}.

\subsection{Summary of Fidelity Metrics}
Assessing the quality of artificial medical images is tricky because there are many ways to measure success. \textbf{Table}~\ref{tab:metrics_summary} summarizes these key metrics, which are essential to determine whether the generated images are realistic and useful for heart segmentation.

\begin{table*}[htbp]
\caption{Summary of Fidelity Metrics Used in CMRI Synthetic Image Evaluation}
\label{tab:metrics_summary}
\begin{center}
\scriptsize
\begin{tabular}{|p{2.6cm}|p{5cm}|p{2.5cm}|p{3.2cm}|p{2.2cm}|}
\hline
\textbf{Metric} & \textbf{Definition} & \textbf{Level (Pixel / High)} & \textbf{Application (Generation / Segmentation)} & \textbf{Frequency in Literature} \\
\hline

Fréchet Inception Distance (FID) &
Measures distribution similarity between real and generated image features; captures fidelity and diversity. &
High (Feature / Distribution) &
Image Generation (Standard metric for GANs / Diffusion models) &
Very High \\
\hline

Structural Similarity Index Measure (SSIM) &
Measures structural similarity and perceptual quality between two images. &
Pixel-Level (Structural) &
Image Generation; MRI Reconstruction &
High \\
\hline

Multi-Scale SSIM (MS-SSIM) &
Multi-scale variant of SSIM assessing structure at various resolutions. &
Pixel / High (Multi-scale) &
Image Generation (Structural fidelity) &
High \\
\hline

Peak Signal-to-Noise Ratio (PSNR) &
Measures pixel-level fidelity based on error magnitude. &
Pixel-Level (Error) &
MRI Reconstruction; Image Generation &
High \\
\hline

Dice Similarity Coefficient (DSC) &
Measures overlap between segmentation prediction and ground truth. &
High (Spatial Overlap) &
Segmentation Models; Utility of Synthetic Data &
Very High \\
\hline

Hausdorff Distance (HD) &
Measures boundary distance between two shapes/segmentations. &
Pixel / High (Boundary Distance) &
Segmentation Models (Boundary accuracy) &
Medium \\
\hline

Kernel Inception Distance (KID) &
Evaluates generative performance using Inception features, unbiased alternative to FID. &
High (Feature / Distribution) &
Image Generation &
Low \\
\hline

Inception Score (IS) &
Measures quality and diversity of generated images using classification confidence. &
High (Feature / Distribution) &
Image Generation &
Low \\
\hline

Fréchet ResNet Distance (FRD) &
Alternative distance metric using ResNet-based features for generative evaluation. &
High (Feature / Distribution) &
Image Generation &
Low \\
\hline

Reconstruction Fréchet Inception Distance (rFID) &
Measures perceptual quality for image reconstruction relative to ground truth. &
High (Feature / Distribution) &
Image Reconstruction / Inpainting &
Low \\
\hline

\end{tabular}
\label{tab:fidelity_metrics}
\end{center}
\end{table*}

\section{Privacy Evaluation in Generative Models}
\subsection{Balancing Privacy and Image Fidelity}
Evaluating privacy in generative cardiac MRI systems requires navigating a core tension: protecting patient identity while preserving clinically useful image fidelity. Differential Privacy (DP) offers formal guarantees by injecting noise to limit the influence of any individual patient \cite{Sun2023}. However, for high-dimensional medical images, strict privacy budgets often degrade anatomical detail, producing blurry or structurally unreliable outputs \cite{Torfi2021}. Because DP tends to conflict with the fine-grained realism required in clinical imaging, recent work increasingly favors attack-based, post-training privacy evaluation, where models are stress-tested like adversaries to identify potential leakage \cite{bergen}.

\subsection{Membership Inference and Frequency-Calibrated Attacks}
A common attack-based technique is the Membership Inference Attack (MIA), which attempts to determine whether a specific real image was part of the training set by analyzing reconstruction error patterns \cite{bergen}. Privacy quantification here is twofold: attack-based metrics measure adversarial success rates, while overfitting-based metrics estimate information leakage by analyzing the generalization gap between training and validation performance \cite{Padariya2025}. However, standard MIAs often perform poorly on medical images due to “inherent image difficulty” -cardiac MRI contains polarized frequency components that make high-frequency details equally challenging for both members and non-members \cite{Zhao2025}.\\
To address this, Frequency-Calibrated Reconstruction Error (FCRE) has been introduced. FCRE isolates mid-frequency components removing noisy high frequencies and uninformative low frequencies to better differentiate true memorization from general reconstruction limitations \cite{Zhao2025}.

\subsection{Structural Similarity Audits and Mode Collapse Checks}
Beyond membership inference, privacy auditing must also examine whether synthetic samples structurally replicate real patients. Pairwise similarity attacks compute the minimum pixel or latent-space distance between synthetic outputs and real images to detect nearest-neighbor leakage \cite{Xing2023}. This risk is amplified in small datasets or rare-pathology settings \cite{Li2025}.\\

Complementary distribution-based attacks check whether many synthetic samples cluster around a single real patient, indicating mode collapse–driven memorization. Such multi-level auditing is especially important in style-conditioned models, where scanner-specific artifacts may act as unintended identifiers \cite{Wong2025}, and in deterministic generative frameworks like flow matching, where fixed noise-to-sample mappings require additional privacy stress tests \cite{Moschetto2025}.

\subsection{Practical Safeguards and Regulatory Alignment}
Regulatory discussions increasingly emphasize that anonymization cannot rely solely on transformations; instead, layered safeguards are required. To standardize these evaluations, recent frameworks propose optimized metrics suites that weigh and compose diverse privacy and utility metrics (e.g., fidelity, specific utility) to rigorously assess trade-offs \cite{Padariya2025}. Recommended measures include:
\begin{enumerate}
    \item \textbf{Technical transformations} - resolution / intensity harmonization and removal of non-essential anatomy \cite{Wong2025},
    
    \item \textbf{Generative training constraints} - DP-enforced optimization \cite{Torfi2021}, diversity maximization to reduce mode collapse-linked memorization \cite{Xing2023},

    \item \textbf{Post-generation audits} - MIAs with anatomically matched conditions \cite{bergen}, nearest-neighbor searches across entire protected archives \cite{Xing2023},

    \item \textbf{Policy controls} - usage contracts forbidding re-identification attempts and outlining permissible applications.
\end{enumerate}

\begin{table*}[htbp]
\centering
\scriptsize
\caption{Summary of Cardiac MRI (CMRI) Datasets Used in Generative Modelling and Segmentation Studies}
\label{tab:datasets_summary}
\renewcommand{\arraystretch}{1.35}
\begin{tabular}{|p{3.3cm}|p{3.4cm}|p{2.2cm}|p{3.2cm}|p{4.1cm}|}
\hline
\textbf{Dataset Name} & \textbf{Vendors / Sites} & \textbf{Number of Samples} & \textbf{Label Availability} & \textbf{Usefulness} \\
\hline

\textbf{M\&Ms Challenge} (Multi-Center, Multi-Vendor, Multi-Disease) &
6 hospitals across Spain, Canada, Germany; Vendors: Siemens, Philips, GE, Canon &
350 images \cite{AlKhalil2023}; 150 annotated, 25 unannotated &
Open-access; LV/RV/MYO labels at ED/ES annotated by clinicians &
Most diverse CMR dataset; benchmark for generalization across vendors \cite{Campello2021}; used in image synthesis and multi-tissue segmentation studies \cite{AlKhalil2023}. \\
\hline

\textbf{UK Biobank (UKB)} &
Single center; Siemens MAGNETOM Aera 1.5T &
Over 5,000 subjects; some studies use 43,352 participants or 25,480 SAX cine sequences &
Restricted access; LV/RV annotations at ED/ES &
Used for large-scale cardiac function studies, phenotype-guided generation \cite{Phenotype2025}, fairness correction, debiasing, and conditional CMR synthesis \cite{Xiangli}. \\
\hline

\textbf{ACDC Challenge} &
Single clinical center &
150 subjects &
Public; LV/RV/MYO segmentation at ED/ES; includes NOR, DCM, HCM pathologies &
Established segmentation benchmark \cite{Campello2021}; widely used for domain shift \cite{Wong2025} and deep learning evaluation. \\
\hline

\textbf{OCMR Dataset} &
Siemens Prisma 3T, Avanto 1.5T, Sola 1.5T &
53 fully sampled + 212 under-sampled (total 1383 k-space entries) &
Open-access; full multi-coil k-space data (healthy subjects) &
Used for DL reconstruction, acceleration studies \cite{Al-Haidri2023}, and unconditional diffusion model training \cite{Strzelecki2025}. \\
\hline

\textbf{M\&Ms-2 Challenge} &
Multi-vendor; testing on Siemens, Philips, GE, Canon &
50 studies per vendor; extra 50 from unseen vendor (Canon) &
Segmentation labels for LV/RV/MYO &
Rigorous benchmark for generalizing to unseen moderate/severe cardiac pathologies \cite{Amirrajab2023}. \\
\hline

\textbf{MS-CMRSeg Challenge} &
Multi-sequence challenge (LGE focus) &
45 subjects &
Manual LV/RV/MYO labels by three observers; LGE-CMR images &
Used for domain transfer, LGE-focused synthesis and segmentation methods \cite{Zhang2025}. \\
\hline

\textbf{MM-WHS Challenge} (Multi-Modality Whole Heart Segmentation) &
Multi-center, multi-modality (MRI + CT) &
120 images (60 CT, 60 MRI) &
Labels for 7 structures (LV, RV, LA, RA, Myo, Ao, PA) &
Benchmark for whole-heart 3D segmentation \cite{Zhang2025}. \\
\hline

\textbf{Sunnybrook Dataset} &
N/A &
45 cases &
Public; LV segmentation labels \cite{Liu2024} &
Used for segmentation algorithm evaluation and comparison. \\
\hline





\end{tabular}
\label{tab:cmri_datasets}
\end{table*}

\section{Utility Evaluation in Generative Models}
\subsection{Segmentation Performance and Fidelity}
The primary measure of synthetic data utility in cardiac MRI is its ability to improve segmentation performance, especially in accurately delineating cardiac structures \cite{AlKhalil2023}\cite{Li2025}. A standard evaluation strategy involves training segmentation networks solely on synthetic datasets generated using GANs, VAEs, or diffusion models and comparing their accuracy with models trained on real images \cite{Diller2020}\cite{Xing2023}. Methods incorporating segmentation-informed conditional GANs have reported measurable gains, including up to a 4\% increase in Dice score and a 40\% reduction in Hausdorff distance when synthetic samples are added to real training data across multi-site datasets \cite{AlKhalil2023}.\\

Mask-conditioned diffusion frameworks further enhance anatomical precision, demonstrating superior myocardium boundary reconstruction relative to unconditioned generators, which often fail to retain endocardial and epicardial detail \cite{Konz2024}.
Notably, diffusion-generated cardiac MRIs achieve left ventricle segmentation Dice scores that closely match those obtained from real data and consistently outperform unconditional GAN-based synthesis \cite{Strzelecki2025}\cite{Li2025}. Results from Progressive GAN studies indicate negligible differences (below 1\%) in segmentation metrics between models trained on real versus synthetic sets, reinforcing the clinical promise of these approaches \cite{Diller2020}.

\subsection{Cross-Vendor Robustness and Dataset Diversity}
Clinical deployment requires models that generalize across MRI scanners, where hardware differences introduce domain shifts affecting contrast and noise characteristics \cite{Amirrajab2023}. Synthetic data helps mitigate this variability through vendor-style conditioned generation, enabling models to learn scanner-specific visual properties without altering underlying anatomical structures \cite{AlKhalil2023}\cite{Wong2025}. However, such style adaptation must be carefully interpreted to ensure that subtle pathology-related features remain detectable \cite{Amirrajab2023}.\\

Synthetic generation also increases dataset diversity by exposing models to a broader spectrum of anatomical and stylistic variations \cite{Moschetto2025}\cite{AlKhalil2023}. which is particularly valuable for downstream phenotype prediction. For instance, cardiac phenotype-guided generative models have improved performance on metrics such as Left Ventricular End-Diastolic Volume (LVEDV) and Left Ventricular Ejection Fraction (LVEF), demonstrating that synthetic data can strengthen clinical outcome prediction tasks \cite{Phenotype2025}.

\subsection{Architectural Trade-offs and Privacy}
Utility varies significantly by generative architecture. GANs can introduce spurious high-frequency artifacts that segmentation networks mistakenly interpret as boundaries \cite{Li2025}. Diffusion models, by contrast, gradually refine images through iterative denoising, yielding smoother textures and more stable volumetric estimations across heterogeneous test datasets \cite{Kebaili2023}.\\

Clinical evaluations further highlight the realism of diffusion-generated cardiac MRI, with radiologists achieving only 60\% accuracy when asked to distinguish synthetic from real images close to chance-level discrimination \cite{Strzelecki2025}.\\

Regarding privacy, integrating differential privacy mechanisms may degrade anatomical quality through noise-induced blurring \cite{Torfi2021}. Therefore, complementary assessments such as membership inference attacks remain necessary to ensure that synthetic samples do not encode identifiable patient-specific features, maintaining low detection accuracy to minimize leakage risk \cite{bergen}\cite{Sun2023}.

\subsection{Augmentation Efficiency and Practical Constraints}
The volume and quality of synthetic augmentation must be carefully balanced. Excessive synthetic samples can reduce performance if the generator collapses toward narrow stylistic modes, reducing dataset diversity \cite{Xing2023}. This issue is especially relevant for rare pathology augmentation, where small but targeted synthetic additions often provide more substantial improvements than large volumes of normal-structure images \cite{AlKhalil2023}\cite{Amirrajab2023}.\\

Computational cost is another constraint: diffusion-based generation remains significantly slower than GANs and VAEs. For example, synthesizing a 128×128 cardiac MRI slice using diffusion takes \textasciitilde 2.31 seconds, compared to 0.12 seconds for GANs and 0.08 seconds for VAEs \cite{Strzelecki2025}. Such disparities influence feasibility for large-scale or real-time clinical pipelines.

\section{Comparative Analysis}
\FloatBarrier
\subsection{Summary of Datasets}
To contextualize the diversity and limitations of existing cardiac MRI resources, we summarize the major datasets used in generative modelling and segmentation studies, highlighting sample size, vendor variability, and available annotations \textbf{Table}~\ref{tab:datasets_summary}.


\subsection{Summary of CMRI Generative Models}
We further provide a consolidated overview of the generative models applied to CMRI synthesis, comparing their conditioning strategies, backbone architectures, and reported performance across fidelity, utility, and privacy dimensions \textbf{Table}~\ref{tab:model_summary}.
\begin{table*}[htbp]
\caption{Summary of CMRI Generative Models from Literature}
\label{tab:model_summary}
\centering
\scriptsize
\renewcommand{\arraystretch}{1.35}
\begin{tabular}{|p{1.2cm}|p{2.5cm}|p{2.5cm}|p{2.5cm}|p{2.4cm}|p{2.4cm}|p{2.4cm}|}
\hline
\textbf{Reference} & \textbf{Generative Backbone} & \textbf{Dataset (Size / Limitation)} 
& \textbf{Conditioning (Mask / Covariate / Other)} & \textbf{Fidelity Evaluation} 
& \textbf{Utility Evaluation} & \textbf{Privacy Evaluation} \\ \hline

\multicolumn{7}{|c|}{\textbf{UNCONDITIONAL GENERATION}} \\ \hline

Bubeck et al. (2022) \cite{bubeck2022} &
UViT-B2 / DiT-B2 (Diffusion), MaskGIT (Autoregressive) &
UK Biobank (11,360 subjects; \textbf{large cohort}) &
Unconditional generation &
FID (38.78), KID &
Benchmarking across spectrum; MaskGIT robust to masking &
Not reported \\ \hline

Özdemir \& Eroğul (2024) \cite{Strzelecki2025} &
DDPM with asymmetric attention-enhanced U-Net &
OCMR dataset (\textbf{limited data}, restricted views) &
Unconditional &
FID (77.78), SSIM, MS-SSIM; radiologist accuracy 60\% &
Improves clinical downstream tasks; anatomical fidelity confirmed &
Not reported \\ \hline

Tetralogy of Fallot Study (PG-GAN) \cite{Diller2020} &
Progressive GAN (PG-GAN), 256$\times$256 output &
Tetralogy of Fallot dataset (303 patients; \textbf{limited}) &
Unconditional latent-based generation &
Expert review; Dice comparison against GT masks &
Synthetic training achieves U-Net accuracy within 1\% of real-data baseline &
Synthetic images free of privacy constraints \\ \hline

\multicolumn{7}{|c|}{\textbf{CONDITIONAL GENERATION}} \\ \hline

Al Khalil / Amirrajab (2022–2023) \cite{AlKhalil2023}\cite{Amirrajab2022} &
Conditional GAN with SPADE layers &
M\&Ms Challenge (multi-vendor/multi-disease; \textbf{data scarcity}) &
\textbf{Multi-tissue segmentation maps} (LV, RV, Myo, lung, fat, muscle) &
SSIM, PSNR, NRMSE; SPADE superior to Pix2Pix/Pix2PixHD &
Up to 4\% Dice improvement; 40\% HD reduction; replacement viability &
Addresses privacy constraints in data sharing \\ \hline

Amirrajab et al. (2023) \cite{Amirrajab2023} &
Two-stage: VAE (deformation), GAN (label-conditioned) &
ACDC + M\&Ms (pathology-limited datasets) &
\textbf{Mask latent space manipulation} for pathology synthesis &
EDV/ESV distribution comparison, Dice, HD &
Enhances generalization across vendors and unseen pathologies &
Not reported \\ \hline

Gheorghiță et al. (2022) \cite{Gheorghi2022} &
Conditional GAN (GauGAN-based mask-to-image) &
Kaggle + UK Biobank (\textbf{scarce, unbalanced}) &
\textbf{Binary masks} with uniform EF distribution &
Clinical metric accuracy focused (not explicit fidelity metrics) &
Pretraining improves regression accuracy for EF/EDV/ESV &
Not reported \\ \hline

Campello et al. (2022) \cite{Campello2021} &
Conditional Generative Adversarial Network (cGAN) &
UK Biobank (43,352 subjects; \textbf{cross-sectional}) &
\textbf{Covariates:} age, BMI, ED/ES cardiac phase &
Qualitative; MAE from pretrained ResNet18 regressor &
Extracts longitudinal patterns; debiases datasets &
Not reported \\ \hline

Cheng et al. (2023) \cite{Cheng2024} &
Two-stage 3D Latent Diffusion (VAE + DDIM) &
UK Biobank (927 sequences; \textbf{large}) &
\textbf{3D segmentation masks} stacked with images &
FID (28.37), FRD (2.92), MS-SSIM, Precision/Recall &
Synthetic images/masks viable as real-data replacements &
Not reported \\ \hline

Li et al. (2024) \cite{Phenotype2025} &
Masked Autoregressive Diffusion (CPGG) &
UK Biobank (32,444 cine), private CMDS dataset (\textbf{large-scale}) &
\textbf{82 cardiac phenotypes} (LVEF, LVEDV, etc.) &
FID, Fréchet Video Distance (FVD) &
Improves downstream disease classification and phenotype prediction &
Not reported \\ \hline

Skorupko et al. (2023) \cite{Skorupko2025} &
Latent Diffusion (Stable Diffusion) + ControlNet &
UK Biobank (\textbf{biased, underrepresented groups}) &
\textbf{Textual attributes} + \textbf{segmentation masks} (shape) &
FID, qualitative analysis of BMI features &
Targeted augmentation reduces algorithmic bias &
Mitigates privacy \& fairness issues \\ \hline

Ma et al. (2024) \cite{Li2025} &
Two-stage Controllable Mask Diffusion (LDM + SPADE) &
M\&Ms Challenge (Canon/Siemens; \textbf{data scarcity}) &
\textbf{Multi-category spatial masks} (LV, RV, septum) &
Dice, IoU (segmentation proxy for fidelity) &
5–10\% DSC improvement over traditional augmentation &
Not reported \\ \hline

Zou et al. (2023) \cite{Zou2023} &
Conditional VAE + Generative SToRM (manifold reconstruction) &
IRB-approved contrast-sequence dataset &
\textbf{Motion signals} + \textbf{inversion time (contrast)} &
Qualitative cine comparison vs bSSFP CINE &
Generates T1 maps + CINE MRI simultaneously; improves efficiency &
Not reported \\ \hline

\end{tabular}
\label{tab:cmri_generative_models}
\end{table*}

\subsection{Summary of CMRI Evaluations}
\begin{figure}
    \centering
    \includegraphics[width=1\linewidth]{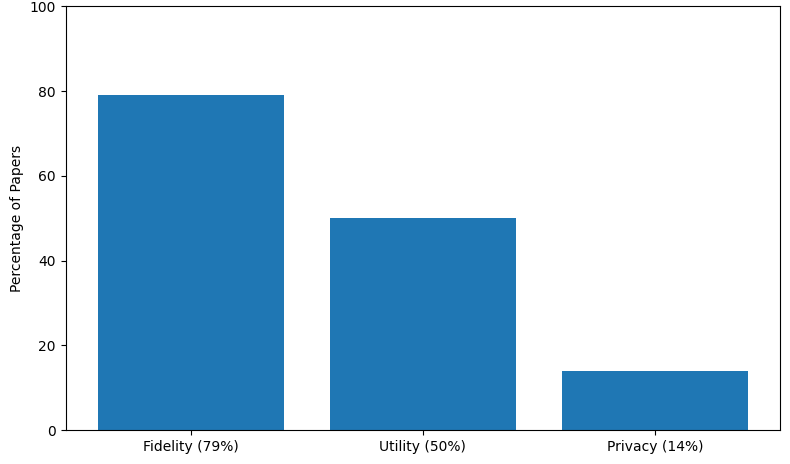}
    \caption{Summary of Fidelity, Utility, and Privacy Reporting}
    \label{fig:evaluationSummary}
\end{figure}

\textbf{Figure}~\ref{fig:evaluationSummary} summarizes the proportion of studies that report fidelity, utility, and privacy evaluations, showing that while fidelity is widely assessed, utility is less consistently reported and privacy evaluation remains largely neglected in current CMRI generative modeling literature.

\section{Research Gaps and Future Directions}
\subsection{What the Literature Agrees On}

Generative model progression:
GANs established early mask-to-image translation capabilities \cite{Amirrajab2022}. Diffusion models now offer state-of-the-art fidelity, stability, and mode coverage \cite{Dhariwal2021, Cheng2024, Kebaili2023}, whereas flow-matching shows promise but overfits on small medical datasets \cite{Moschetto2025}. SPADE-based conditioning achieves strong fidelity–efficiency trade-offs for cardiac MRI \cite{Li2025, Amirrajab2022}.

Value of mask conditioning:
Studies consistently show that multi-structure anatomical masks outperform cavity-only labels, improving realism, shape diversity, and structural accuracy \cite{Amirrajab2022}. High-quality semantic guidance often yields greater gains than simply increasing dataset size \cite{Amirrajab2022}.

Utility of pathology synthesis:
Although underexplored, incorporating synthetic pathological variations (e.g., altered geometries indicating hypertrophy or dilation) significantly improves segmentation robustness for rare diseases \cite{Amirrajab2023}.

Lack of privacy evaluation:
Despite motivations rooted in data confidentiality, most CMRI synthesis studies omit quantitative privacy assessment, including membership inference or nearest-neighbor leakage tests.

Insufficient cross-dataset evaluation:
Most works remain limited to within-dataset validation, and the few cross-dataset evaluations (e.g., ACDC → M\&Ms) report substantial performance drops \cite{AlKhalil2023}.

\subsection{Identified Research Gaps}
\subsubsection{Limited pathology-specific mask-conditioned generation}
Existing models primarily synthesize healthy or generic cardiac shapes. Missing capabilities include controlled synthesis of hypertrophy, DCM, ischemic defects, and congenital abnormalities \cite{Amirrajab2023, Amirrajab2022}. \\
Future direction: extend latent-space pathology modelling, enable continuous severity control, and validate realism through expert clinical review.

\subsubsection{Underexplored diffusion and flow-matching performance on small CMRI datasets}
Flow-matching models overfit under small-sample regimes \cite{Moschetto2025}, and diffusion inference remains slow ($\approx 40$s per sample). \\
Future direction: systematically evaluate these architectures under limited-data conditions, develop accelerated cardiac-specific variants, and compare all models under identical fidelity–utility–privacy protocols.

\subsubsection{Absence of systematic privacy evaluation}
Cardiac MRI generative studies rarely include quantitative privacy assessment \cite{Amirrajab2022, Cheng2024}. \\
Future direction: integrate MIAs with calibrated thresholds, differential privacy budgets, and privacy–utility trade-off curves. Establish minimum privacy reporting standards for medical generative models.

\subsubsection{Insufficient cross-dataset utility evaluation}
Most segmentation utility tests rely on single datasets and lack pathology-stratified or vendor-aware validation \cite{AlKhalil2023}. \\
Future direction: adopt standardized cross-dataset protocols (e.g., ACDC → M\&Ms), evaluate by pathology subtype, and use fixed segmentation backbones for reproducible comparisons.

\section{Conclusion}
    Synthetic cardiac MRI generation has emerged as a powerful strategy to mitigate data scarcity, vendor variability, and privacy restrictions that limit real-world medical imaging datasets. Contemporary generative models, particularly mask-conditioned diffusion frameworks, demonstrate strong capability in producing anatomically faithful MRIs that enhance downstream segmentation accuracy and support rare-pathology augmentation. These models outperform unconditional generators by enforcing structural fidelity through segmentation-guided synthesis while preserving fine-grained cardiac morphology. However, achieving robust generalization remains challenging due to multi-vendor acquisition differences, motivating the use of vendor-style conditioning and harmonization to maintain anatomical integrity across domains. Privacy remains a critical concern, with risks of memorization and nearest-neighbor leakage requiring comprehensive audits, including membership inference, structural similarity checks, and differential privacy-aware training. Current research gaps, such as limited pathology-specific conditioning, inadequate multi-dimensional privacy evaluation, and insufficient head-to-head model comparisons highlight the need for integrated pipelines that jointly optimize fidelity, diversity, and privacy for clinically safe synthetic cardiac MRI generation.

\clearpage

\section*{Glossary Terms}

\subsection*{Imaging \& Anatomy Abbreviations}
\textbf{CMR / CMRI} — Cardiac Magnetic Resonance Imaging.

\textbf{MRI} — Magnetic Resonance Imaging

\textbf{LV} — Left Ventricle

\textbf{RV} — Right Ventricle

\textbf{MYO} — Myocardium

\textbf{ED} — End-Diastolic phase

\textbf{ES} — End-Systolic phase

\textbf{SAX} — Short-Axis view

\textbf{FOV} — Field of View (extent of MRI acquisition)

\textbf{ROI} — Region of Interest (cropped region containing the heart)

\subsection*{Generative Modelling Terms}
\textbf{GAN} — Generative Adversarial Network

\textbf{VAE} — Variational Autoencoder

\textbf{Diffusion Model} — A generative model trained to iteratively denoise samples

\textbf{DDPM} — Denoising Diffusion Probabilistic Model (canonical diffusion model)

\textbf{Forward Markov Process} — The noise-adding process in diffusion models that gradually destroys image structure

\textbf{Flow-Matching Model} — A model learning continuous, deterministic transformations between distributions

\textbf{Mask-Conditioned Generation} — Using segmentation masks as structural guidance during image synthesis

\textbf{Latent Space} — Compressed representation where models encode high-level features

\textbf{de facto} — A standard method or practice that is widely used in practice, though not formally designated

\textbf{MONAI} — Medical Open Network for AI; an open-source framework for training medical imaging models

\subsection*{Evaluation Metrics}
\textbf{MAE} — Mean Absolute Error

\textbf{MSE} — Mean Squared Error

\textbf{SSIM} — Structural Similarity Index Measure

\textbf{FID} — Fréchet Inception Distance (measures realism of generated images)

\textbf{FCRE} — Feature Consistency and Reconstruction Error (used in some generative evaluations)

\subsection*{Privacy-Related Terms}
\textbf{MIA} — Membership Inference Attack (tests whether model reveals if a sample was in training data)

\textbf{Nearest-Neighbour Leakage} — Privacy risk where generated data resembles real training samples

\textbf{Differential Privacy (DP)} — Formal framework adding controlled noise to ensure privacy guarantees

\subsection*{Data \& Domain Shift Terms}
\textbf{Vendor Variability} — Differences in scanners, protocols, reconstruction methods across manufacturers

\textbf{Domain Shift} — Performance drop when model is tested on data from distributions different from training

\textbf{Cross-Vendor Generalization} — Ability of a model to perform consistently across different scanner vendors






\subsection*{Dataset Names}
\textbf{M\&Ms Dataset} — Multi-Center, Multi-Vendor, Multi-Disease cardiac MRI dataset

\textbf{ACDC Dataset} — Automated Cardiac Diagnosis Challenge dataset

\bibliographystyle{IEEEtran}
\bibliography{export}

\end{document}